\documentclass[letterpaper,twocolumn,fleqn,11pt]{article}

\usepackage{ist}
\usepackage{multirow}
\usepackage{tabularx}
\usepackage{mathtools}
\usepackage{hyperref}
\usepackage{anyfontsize}
\title{Extreme Face Inpainting with Sketch-Guided Conditional GAN}

\author{Nilesh Pandey and Andreas Savakis; Rochester Institute of Technology; Rochester, NY, USA.}

\begin{document} 
% \makeatletter
% \providecommand*{\toclevel@unnumberedsection}{0}
% \makeatother
\maketitle 

% \thispagestyle{empty} % prevents the first page to be numbered

%%%%%%%%%%%%%%%%%%%%%%%%%%%%%%%%%%
% Abstract
%%%%%%%%%%%%%%%%%%%%%%%%%%%%%%%%%%
\raggedbottom
\begin{abstract}
Recovering badly damaged face images is a useful yet challenging task, especially in extreme cases where the masked or damaged region is very large. One of the major challenges is the ability of the system to generalize on faces outside the training dataset. We propose to tackle this extreme inpainting task with a conditional Generative Adversarial Network (GAN) that utilizes structural information, such as edges, as a prior condition. Edge information can be obtained from the partially masked image and a structurally similar image or a hand drawing. In our proposed conditional GAN, we pass the conditional input in every layer of the encoder while maintaining consistency in the distributions between the learned weights and the incoming conditional input. We demonstrate the effectiveness of our method with badly damaged face examples.
% Tasks like image recovery, and image editing are very challenging topics in computer vision.
% Photoshop, a widely popular tool used for image editing cannot always generate realistic images, if the editing region is too large.
% Image recovery is trivial task if the damaged region is small enough to approximate, but a difficult problem if the damaged region is large, since we have no prior information on the subject.
% We tackle image recovery problem by extracting structural information like edges from the partially masked image or from a structurally similar image that can be used as a prior condition.
% Additionally, editing the edges of the partially destroyed region can be used as a prior condition.
% We use an existing GAN (Generative Adversarial Network) and propose novel training method for image inpainting and image editing.
% We perform subsequent experiments to show the significance of our proposed method.
% In our proposed method, we pass the conditional input in every layer of the encoder while maintaining the relative statistics between the learned weights and the incoming conditional input. The main aim of the paper is to generalize inpainting tasks for different applications including image inpainting and image editing that many recent state-of-the-art methods are unable to perform. We believe that for a method to be useful in real-world applications, it should be able to generalize across the images which are different from the datasets they are trained on.

\end{abstract}

\section{Introduction}
Image inpainting is a popular and challenging problem in computer vision with applications in image editing and restoring of damaged image regions \cite{ipgan}, \cite{partial}, \cite{gatedconv}, \cite{multi}, \cite{scfegan}, \cite{faceshop}. 
The damaged image regions may range from rectangular patches of various sizes to irregular regions or entire objects.
In image editing, some regions are tactically damaged so that they can be inpainted with different content.

\begin{figure}
    \centering
    \includegraphics[width=\columnwidth]{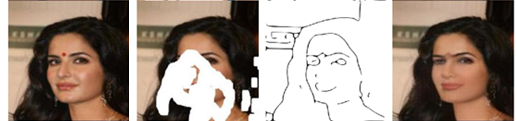}
    \caption{Example of sketch-conditioned face inpainting with the proposed FIpoly-GAN network. From left to right: original, damaged image with large irregular mask, hand-drawn sketch, inpainted result. }
    \label{face_example}
\end{figure}

In this paper we deal with the problem of extreme inpainting, where large regions of faces have been damaged with the application of an irregular mask. Initial works dealt with small square patches, while more recent approaches take on the more challenging problem of inpainting large irregular regions \cite{gatedconv}, \cite{multi}. 
Our face inpainting method, called FIpoly-GAN, is based on the Poly-GAN conditional Generative Adversarial Network (GAN) architecture \cite{polygan}, that was initially proposed for fashion synthesis
and is suitable for various tasks, e.g. geometric transformations, image stitching and inpainting. To recover the facial structure that is lost after significant image regions have been masked, we utilize an edge sketch as a condition that guides the inpainting process by providing structure in the damaged regions.
The proposed FIpoly-GAN has shown ability to generalize on faces obtained in the wild from various sources unrelated to the training data.
A representative result is shown in in Fig. \ref{face_example}.  
% To create a sense of generalization we try to test existing state-of-the-art \cite{multi} results on an empty image consisting of a sky. We observe that most of the methods tend to create a face (Figure:\ref{fig:false_face}) in the empty region. We use this premise as a motivation for our research in an attempt to find an algorithm or technique that can help us to avoid such scenarios. We will discuss this in detail in upcoming sections. We will discuss the importance of guided information in any Image Inpainting, and Image reconstruction. 
% We also try to explore the importance of priors, or first training the model with complete and undamaged image (not damaged) to help the model to converge, and after certain epochs, we introduce the damaged images which helps the model to learn to perform task on the 
The main contributions of this paper can be summarized as follows:
% \vspace{-0.1in}
\begin{itemize}
\item We propose FIpoly-GAN, a conditional GAN architecture that leverages edge information for extreme face inpainting.
\item Our FIpoly-GAN architecture uses the edge map, from a sketch or a similar image, as a condition that is fed at various levels of the network to learn the structure of the image for inpainting color and texture in the missing region. 
% The edge condition is input at multiple levels of the GAN encoder and is enforced through skip connections between the encoder and decoder. 
\item FIpoly-GAN achieves state of the art results on standard face datasets and outperforms other methods on images that are unrelated to the training dataset.
\end{itemize}

\begin{figure*}[th]
\begin{center}
\includegraphics[width=0.95\linewidth]{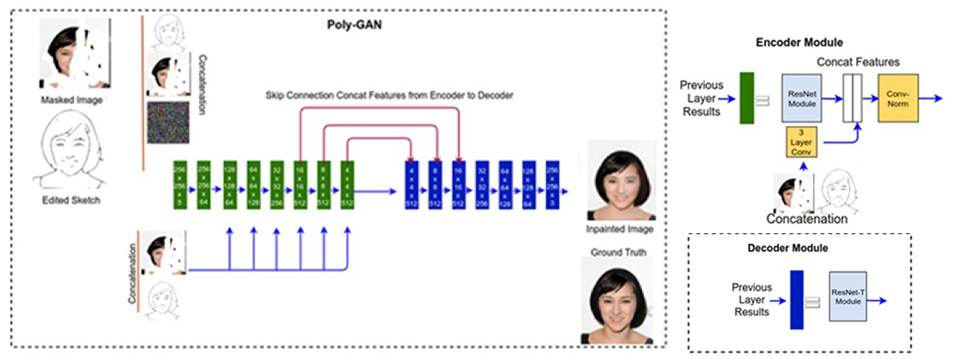}
\end{center}
  \caption{The FIpoly-GAN architecture used for edge-guided face inpainting. The left side block shows the overall architecture and the right side shows more details of the encoder and decoder layers.} 
\label{fig:PolyGANarch}
\end{figure*}

% \begin{figure}
% \centering
%   \includegraphics[width=\columnwidth]{inpainting/false_face_sky.png}
%   \caption{We tried some existing state-of-the-art methods on an empty image, and the result is a clearly average face from models trained on CelebA \cite{celebA}. We use method from the paper~\cite{multi}.}
%   \label{fig:false_face}
% \end{figure}

%%%%%%%%%%%%%%%%%%%%%%%%%%%%%%%%%%%%
% Overall Document Guidelines: Head
%%%%%%%%%%%%%%%%%%%%%%%%%%%%%%%%%%%%
\section{Related Work}
\label{sec:intro}
% \par Many different methods have been proposed for image inpainting. 
Early inpainting methods operated under the assumption that the damaged region is a standard shape, i.e. square or rectangle.
Free form inpainting, where the damaged regions are large and irregular, poses a challenge to many methods due to lack of relevant information that can be used from nearby regions.
A  popular approach is the Patch-based image inpainting network \cite{ipgan}, which progressively learns to associate nearby image contect to the missing region. However, this approach is not as effective in the case of free form inpainting. 

Image Inpainting using Partial Convolution \cite{partial} is one of the first methods to tackle free form inpainting on images with irregular damage. It utilizes a neural network with a partial convolution layer and performs an automatic update on the binary mask.
Given a binary mask, the partial convolution utilizes non-damaged information near damaged region to estimate missing content. 
The next step in the method is to update the binary mask according to the output of the partial convolution, so that if the output is positive for the damaged region then the mask is updated accordingly.

Free-form image inpainting with Gated Convolution (SN-patchGAN) ~\cite{gatedconv} is a state-of-the-art method for inpainting images with irregular damage.
This method introduces a new convolution layer named gated convolution.
The input features in the model are first used to compute the gating value,
$g = sigmoid(W_{g} * x)$, where $x$ denotes the input features.
The final output of the layer is multiplication of learned features and gating values, $y = \phi(W*x) \odot g$. 
The architecture includes a contextual attention module with the refinement network, with both having a gated convolution layer. 
Gated Convolution has inspired several other methods for image inpainting.

Image inpainting via Generative Multi-column Convolutional Neural Networks (GMCNN) \cite{multi} The GMCNN network consists of three sub-networks, one  generator, one global and one local discriminator for adversarial training. 
% A pre-trained VGG network is used for the calculation of Implicit diversified Markov Random Field (ID-MRF) loss. 
Only the generator is deployed when testing on a new image.

SC-FEGAN~\cite{scfegan} and Face-Shop~\cite{faceshop} are state-of-the-art image inpainting methods based on image editing. 
They work by providing the network a simple sketch of the face. 
Both methods use a similar approach to generate sketch images of the face, which can be edited and passed to the model as conditional input. The EdgeConnect method \cite{EdgeConnect}  uses an edge generator to estimate structural information in the missing regions.

The SC-FEGAN trains both the generator and discriminator to generate a  (512x512) image. 
The SC-FEGAN generator uses gated convolution in all layers, except the initial input layer in the encoder and final output layer in the decoder, followed by a local signal normalization layer, as in \cite{PGAN}. 
% The SC-FEGAN network's generator utilizes gated convolution and dilated convolution, and 
The SC-FEGAN discriminator 
% used in SC-FEGAN 
is similar to that in SN-patchGAN ~\cite{gatedconv}. 
% Both models use similar U-net style architecture for the Generator part. 

%%%%%%%%%%%%%%%%%%%%%%%%%%%%%%%%%%
% Graphics and Equations
%%%%%%%%%%%%%%%%%%%%%%%%%%%%%%%%%%

\section{Proposed Inpainting GAN Architecture}

\subsection{Generator}
In the proposed edge-guided inpainting approach, we use the FIpoly-GAN architecture shown in Figure~\ref{fig:PolyGANarch}, which is based on the Poly-GAN network initially used for fashion synthesis ~\cite{polygan}. The main differences between the FIpoly-GAN inpainting architecture and the original Poly-GAN architecture
% used for image warping and stitching \cite{polygan} 
are due to different inputs and training methodology. 
% : (a) We change the input to the network. (b) We train the model differently that helps the architecture to perform Image inpainting. 
FIpoly-GAN is a conditional GAN based on encoder-decoder network configuration with two additional modules: the Conv-Module and the Conv-Norm Module, illustrated in Fig. \ref{fig:PolyGANarch}.
We present the architecture, training methodology and results in the following sections.

\subsubsection{Encoder}
The FIpoly-GAN encoder is an extension of deep learning conditional GAN encoders with some variations in the network.
The most important aspect of the FIpoly-GAN encoder is that the conditional input is fed to multiple layers. 
We found that if the conditional input is provided only in the first layer of the encoder, then its effect diminishes as the features propagate deeper in the network. 

In our encoder, each layer is like a ResNet module \cite{Resnet} consisting of two spectral normalization layers followed by a ReLU activation function. 
The encoder in Figure \ref{fig:PolyGANarch} has modules to handle the incoming conditional inputs in every layer using a Conv Module and a Conv-Norm Module, shown in Fig. \ref{fig:PolyGANarch}. % \textbf{Conv Module.} 
In the \textbf{Conv Module}, we pass the conditional inputs to every convolution layer starting with the third layer, followed by a ReLU activation function. 
The learned features from the previous layer have different statistics than the learned features from Conv Module, so we concatenate the features for input through the Conv-Norm Module. 
The Conv-Norm Module consists of two convolutions each followed by an instance normalization and an activation function. 
It is important to note that the initial input to the encoder consists of conditional input with Gaussian noise. 

\subsubsection{Decoder}
The FIpoly-GAN decoder, shown in Figure \ref{fig:PolyGANarch} is the same as the one in Poly-GAN \cite{polygan}.
The layers in the decoder consist of a ResNet  module with two spectral normalization layers followed by a ReLU and a transposed convolution to upsample the learned features.
% This decoder module is similar to the decoders of other GAN's. 
The decoder learns to generate the desired output using the input from the encoder. To increase the effectiveness of the decoder, we utilize skip connections from low resolution layers of the encoder to the corresponding resolution layers of the decoder, as illustrated in Fig. \ref{fig:PolyGANarch}.
% Few layers in the decoder receive extra information from the encoder through the skip connection.
It is known from previous works, such as Progressive GAN \cite{PGAN}, that the coarse layers learn structural information which is easy to manipulate through the conditional input. 
Our experiments determined that only three low resolution skip connections are needed. These skip connections help the decoder enforce the structure of the conditional inputs fed to consecutive layers of the encoder.

% \subsubsection{Skip Connections}
% Poly-GAN utilizes skip connections to pass information from encoder to decoder.
% We know from previous works, such as Progressive GAN \cite{PGAN}, that the coarse layers learn structural information which is easy to manipulate through the conditional input. 
% Similar to the approach in Poly-GAN \cite{polygan} We also test several experiments connecting different layers in the encoder with the layers in the decoder. 
% From experiments, some architectures, with skip connections connecting all layers from the encoder to the decoder, performed equally well in image inpainting tasks but failed in tasks like person re-identification and fashion synthesis, that require a higher degree of structural manipulation. 

\begin{figure}{h}
    \centering
    \includegraphics[width=\columnwidth]{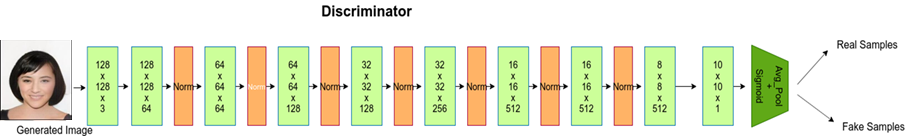}
    \caption{Discriminator network using in FIpoly-GAN. }
    \label{fig:discriminator}
\end{figure}

\subsection{Discriminator}
The FIpoly-GAN discriminator network shown in Figure \ref{fig:discriminator} is the same as the one in \cite{polygan}. The discriminator architecture is similar to the one in Cycle GAN \cite{CycleGAN}, as it has been found to be effective in various applications.

\subsection{Loss}
The loss function used for training consists of three components: adversarial loss $L_{adv}$, GAN loss $L_{gan}$ and identity loss $L_{id}$. 
The total loss and its components are presented below, where $D$ is the discriminator, $G$ is the generator, $x_{i}, i=1,...,N$ represent $N$ input samples from different distributions $p_{i}(x_{i}), i=1,...N$, $t$ is the target, $F$ are fake labels for generated images and $R$ are real labels for ground truth images.
\begin{equation}
\begin{split}
    \underset{G}{\mathit{min}}\mathit{L_{GAN}} = {} & \lambda_{3}E_{x_{1}\sim p_{1}(x_{1}),..,x_{N}\sim p_{N}(x_{N})} \\
    & {\left \| D(G(x_{1},..,x_{N})-R \right \|_{2}^{2}}
    \label{eq:LossGAN}
    \end{split}
\end{equation}

\begin{equation}
\begin{split}
    \underset{D}{\mathit{min}}\mathit{L_{Adv}} = {} & \lambda_{1}E_{t\sim p_{d}(t)}{\left \| D(t)-R \right \|_{2}^{2}} + \\
    & \lambda_{2}E_{x_{1}\sim p_{1}(x_{1}),..,x_{N}\sim p_{N}(x_{N})} \\
    & {\left \| D(G(x_{1},..,x_{N})-F \right \|_{2}^{2}}
    \label{eq:LossAdv}
\end{split}
\end{equation}

% The identity loss is:
\begin{equation}
\begin{split}
    L_{Id}= {} & \lambda_{4}E_{t\sim p_{d}(t),x_{1}\sim p_{1}(x_{1}),..,x_{N}\sim p_{N}(x_{N})}\\
    & {\left \| G(x_{1},..,x_{N})-t \right \|_{1}}
    \label{eq:LossId}
    \end{split}
\end{equation}
In the above equations, 
% \ref{eq:1} to \ref{eq:3}     
 $\lambda_{1}$ to $\lambda_{4}$ are hyperparamters that are tuned during training.
Similarly to ~\cite{LSGAN}, we use the $L_{2}$ loss as the adversarial loss of our GAN in Eqs. (\ref{eq:LossGAN}) and (\ref{eq:LossAdv}).
We use the $L_{1}$ loss for the identity loss, which helps reduce texture and color shift between the generated image and the ground truth. 
We considered adding other loss functions, such as the perceptual loss ~\cite{Perceptual-loss} and SSIM loss, but they did not improve our results.  

For each epoch, we first train the generator with the GAN loss and identity loss in Eqs. (\ref{eq:LossGAN}) and (\ref{eq:LossId}) respectively.  
Then we train the discriminator with the adversarial loss in Eq. (\ref{eq:LossAdv}). 
This process was repeated for several epochs until a satisfactory result was obtained.
% \begin{figure}[htb]
%   \centering
%   \includegraphics[width=\columnwidth, scale=0.20]{inpainting/Gated_Conv.png}
% \caption{We use available online demo for the method using Gated Convolution~\cite{gatedconv}. We lacked capability to test our own images, but it can be seen that the results suffer from visible artifacts.}
% \label{fig:gated_conv}
% \end{figure}

\section{Experiments and Results}
% Due to the attention given to inpainting, many methods have been proposed in recent years and some of them have achieved state-of-the-art on many different datasets. 
% The other important reason for the popularity of the task is due to the availability and release of public datasets like CelebA~\cite{celebA}, Places2~\cite{places}, Labelled Faces in the Wild (LFW)~\cite{LFWTech}. 
% We believe that, to be used in practice, an inpainting method should be able to handle images that it has never seen during training, but may be similar to known images.
% Generally, state-of-the-art methods can generate highly realistic images but a blurry image with artifact when the method encounters image which is too different than the data on which the method was trained.

For our experiments, we used standard datasets, specifically Labeled Faces in the Wild (LFW) \cite{LFWTech} and Large-scale CelebFaces Attributes (CelebA) \cite{celebA}.
% , and Flickr-Faces-HQ Dataset (FFHQ) \cite{StyleGAN} for training. % The datasets have aligned faces in each other respective. 
% The Labeled Faces in the Wild (LFW) website states about LFW:``database of face photographs designed for studying the problem of unconstrained face recognition. 
The LFW dataset contains more than 13,000 images of faces collected from the web. 
Each face is labeled with the name of the person pictured and 1680 people have two or more distinct photos in the dataset. 
% The only constraint on these faces is that they were detected by the Viola-Jones face detector." 
The version of LFW dataset we uses has been aligned using the deep funneling method. 
The CelebA dataset has more than 200k celebrities faces in different scenes, poses, and with and without jewelry. The dataset is pre-processed and aligned. 
% The FFHQ dataset~\cite{StyleGAN} has 70k high-quality images of 1024 resolution maintained and was released by Nvidia Labs. 
% FFHQ has not been used to showcase any work in this paper, but we did experiment during the period of work. 
A key similarity between the datasets is the alignment of facial features, e.g. all images are aligned so that the tip of the nose is at the center of the image. 
Training of FIpoly-GAN was done using free form irregular masks. 

% We first illustrate the shortcomings of existing methods, and then we utilize our approach for handling such scenarios.
% Some of the methods have their code in public release, while others have provided an online demo for their method. 
We  used publicly released code or online demos for comparisons with other methods. 
% In case of unavailability, we tested other methods using the online demo provided by the authors.
We experimented with image inpainting for irregular holes using the method of image inpainting via Generative Multi-column Convolutional Neural Networks ~\cite{multi}, image inpainting Using Gated Convolution ~\cite{gatedconv}, and image inpainting using SC-FEGAN ~\cite{scfegan}, due to the availability of code or online demo.

\begin{figure}[!ht]
  \centering
  \includegraphics[width=\columnwidth]{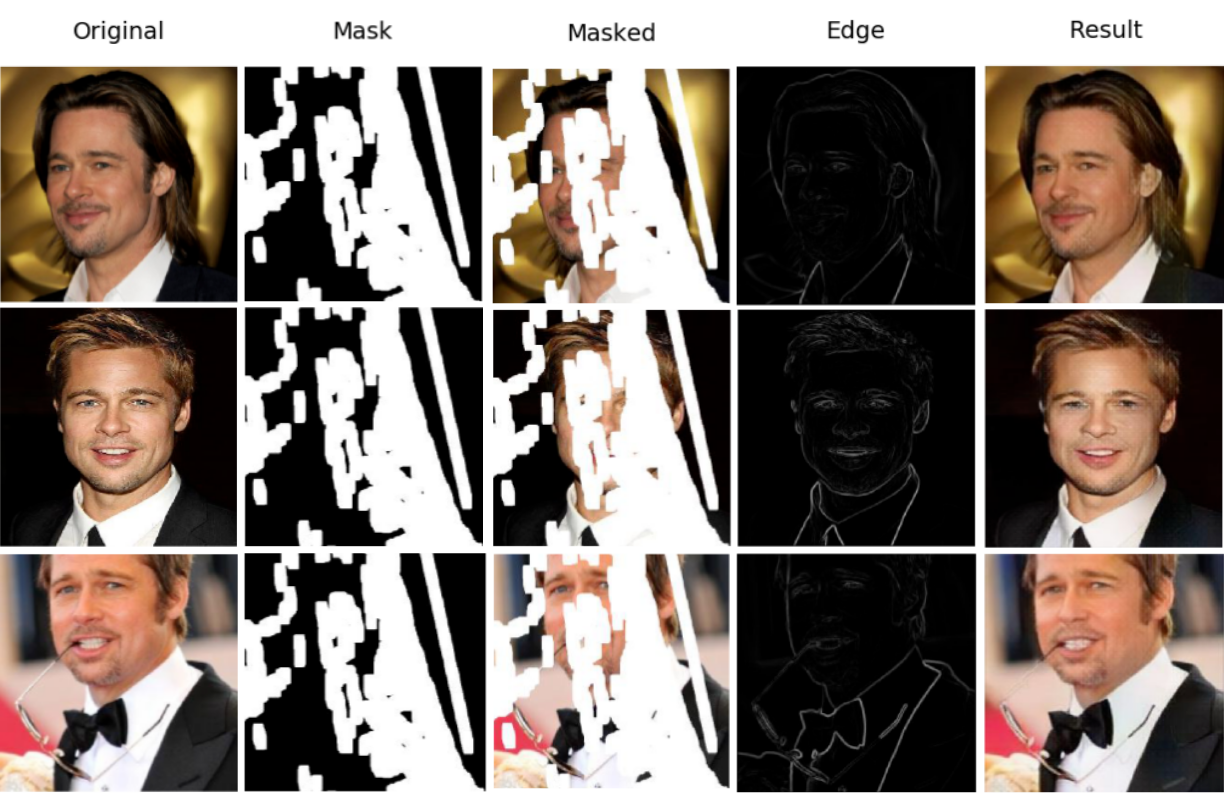}
\caption{Results inpainting with  FIpoly-GAN when passing the edge map as conditional input. These examples use images from the internet to illustrate that FIpoly-GAN is able to generalize to unknown images from random sources.}
\label{fig:edge_based_method}
\end{figure}

\begin{figure}[!ht]
  \centering
  \includegraphics[width=\columnwidth]{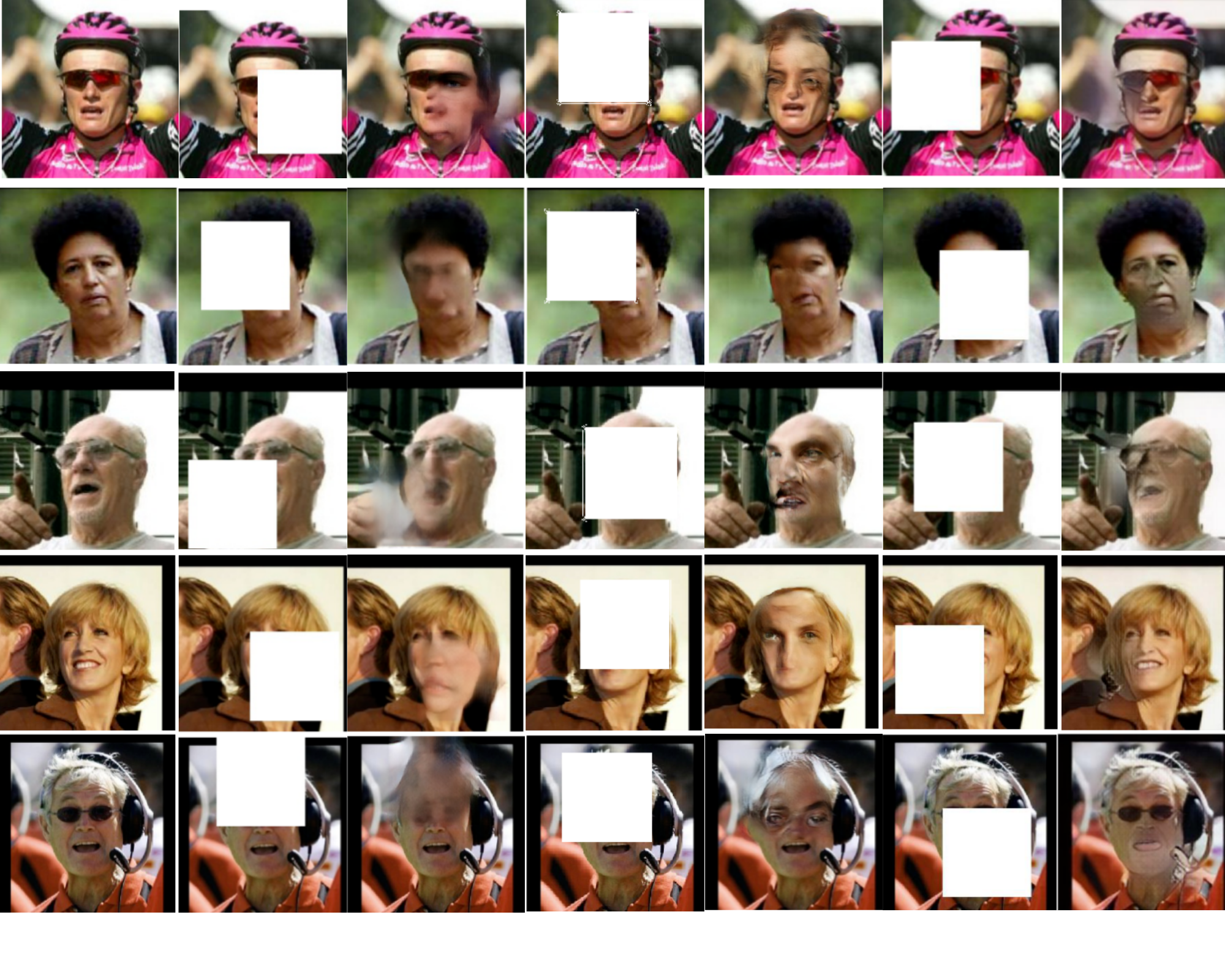}
\caption{Results comparison of edge-based FIpoly-GAN with Generative Multi-column convolutinal nets (GMCNN) \cite{multi} and Gated Convolution method \cite{gatedconv}. 
Left to right columns show: Original, masked, GMCNN result, masked, GatedConv result, masked, edge-based FIpoly-GAN result. Best viewed in color and 4$\times$ zoom.
}
\label{fig:comp_edge_inpainting}
\end{figure}

We next present our experiments with conditional inputs to our network, passing the structural information such as edges, edges of similar-looking images, and edited edges.

\subsection{Edge-based Face Inpainting}

% Most of the mentioned methods output blurry and artifact images when the model comes across the images or similar images it has never seen. One can argue to train the model with more data since we have methods to generate realism in synthetic faces, but the training requires resources. In contrast to feeding more data to the network, passing structural information in the network with the masked images helps to overcome the limitation of the existing methods.

% Most of the mentioned methods output blurry and artifact images when the model comes across the images or similar images it has never seen. One can argue to train the model with more data since we have methods to generate realism synthetic face, but the training requires resources. In contrast to feeding more data to the network, passing structural information in the network with the masked images helps to overcome the limitation of the existing methods. We surround or experiment around passing the structural information such as edges, edges of similar looking images, and edited edges.

Edge-based face inpainting operates under the assumption that the edges of the masked regions are available. In our experiments we obtained the edge map from the original image, operating under a best case scenario.  Alternatively, the edges can be obtained from a similar image or a sketch drawing.
% We will relax this assumption in the upcoming experiments. 
We utilized the processing pipeline as an encoder-decoder style conditional GAN illustrated in Figure \ref{fig:PolyGANarch}. 
For the training data, we randomly sampled nearly 30 thousand images from the CelebA ~\cite{celebA} dataset. We created edge maps for the sampled images using the Sobel filter and thresholded the filter response to obtain the edges that are used as conditions. 
We used an NVIDIA 1080TI GPU for training.
% to train with 30,000 randomly sampled CelebA faces. 
% \subsubsection{Results}

We display representative results from our edge-based inpainting in Figure \ref{fig:edge_based_method} and comparisons with other methods in Figure \ref{fig:comp_edge_inpainting}. Our method is trained using free-form irregular masks on CelebA, while GMCNN is pre-trained on \textbf{rectangular} shaped mask on CelebA. The test images are randomly sampled from the LFW dataset. 
From the results of Fig.~\ref{fig:comp_edge_inpainting} it is evident that our edge-conditioned FIpoly-GAN significantly outperforms the other methods when using images from unknown sources.
In contrast to the other methods, our network has the ability to generalize and recover the masked image content consistently close to the original image. 
% The FIpoly-GAN results are is very close to the original.

% In a real situation, there is very little chance to have the complete edge map of a masked image. 
% As an alternative, one could also pass a detailed sketch of the face as a conditional input to the network.

% Results of example images are displayed in Figure \ref{fig:comp_edge_inpainting}. 
For quantitative comparison with other methods, we trained on the LFW dataset. We used three quantitative metrics to compare our results against state of the art methods: i) Structural Similarity Index (SSIM); ii) Mean Squared Error (MSE); iii) Signal to Noise Ratio (SNR).  
SSIM should be high for similar images, while MSE should be low for similar images. Higher SNR values represent better quality of image recovery.  The results indicate that the edge-based FIpoly-GAN scores highest in two of the metrics. Most importantly, our method is able to generalize much better than the other methods on previously unseen images, as illustrated in Fig.~\ref{fig:comp_edge_inpainting}.

\begin{table}

\begin{center}
\begin{tabular}{||c c c c|| }
 \hline
  Method &  SSIM &  MSE & SNR \\
 \hline
  Edge-based FIpoly-GAN  &  0.824  &  \textbf{27.533} &  \textbf{1.678}  \\
\hline
  GMCNN \cite{multi} &  \textbf{0.845} &  36.041 &  1.568 \\
 \hline
  Gated Conv \cite{gatedconv} &  0.834 &  37.261 &  1.540 \\
 \hline
 
\end{tabular}
\caption{Quantitative results comparison of edge-based FIpoly-GAN with other methods on randomly sampled images from the LFW dataset. Bold shows best performance.}
\label{tab:res}
\end{center}
\end{table}

% \subsection{Reference Based Image Editing}
% \begin{figure}[htb]
%   \centering
%     \includegraphics[width=\columnwidth]{inpainting/Ref_based_inpainting.png}
%   \caption{Results from our approach of passing edge of the reference image as conditional input. We do align the edges of the reference image, basically the same approach as the first experiment.}
%   \label{fig:ref_based_method}
% \end{figure}
% We relax our previous assumption of having the edges of the masked image.
% Instead, we assume having a structurally similar aligned reference image, so we can use the edge of the reference image for the task of image completion of the masked image. 
% For this task, we use the LFW dataset ~\cite{LFWTech} 
% and create aligned reference image used for training and testing.

% % \subsubsection{Results}
% The initial results in Figure \ref{fig:ref_based_method} are not so good, but the important thing is that the masked image can blend the edges of the reference image with the masked image seamlessly. 
% The texture of the source image is maintained in the final image, strengthening the argument that edge information can be useful in image inpainting. 

% \begin{figure*}[hbt!]
% \centering
%   \includegraphics[keepaspectratio=true,scale=1.25]{inpainting/ref_one_mask.png}
%   \caption{Results from edges of reference based inpainting.}
%   \label{fig:ref_inpainting}
% \end{figure*}

\subsection{Sketch-based Face Inpainting}

% From our previous experiments, and from the results of the existing methods in Figures  \ref{fig:gated_conv}, \ref{fig:multi_column}, \ref{fig:sc_fegan}, we conclude that to have an inpainting approach that generalizes for previously unseen images during training, we need to pass more information about the image to the network. 
Our edge-based inpainting experiments demonstrated that even if the image is badly damaged, we can recover the original content by conditioning on the edge information. However, edge information is not readily available when inpainting. 
% The recovered image is very similar to the masked image in the case of Figure \ref{fig:edge_based_method}. 
% In the second scenario, we pass the structural information of a structurally similar image as conditional input to the network.
% , and the results are not too far off the ground truth. 
% From the second scenario of Figure \ref{fig:ref_inpainting}, we conclude that having similar structural information  helps to complete the image almost as well as having the edge map. 
As an alternative, we consider a hand-drawn sketch that provides edge information for the damaged image regions.
We created a dataset consisting of sketches of CelebA faces following the instructions provided in FaceShop ~\cite{faceshop}.
We used Potrace to generate a vectorized representation of the image edges, which is used for training purposes. We randomly sampled 150k images for training and testing purposes from CelebA. The model was trained on Ubuntu 16.04 with an NVIDIA  1080TI GPU for nearly 18 hours.

% \subsubsection{Results}

\begin{figure}[!ht]
  \centering
  \includegraphics[width=\columnwidth]{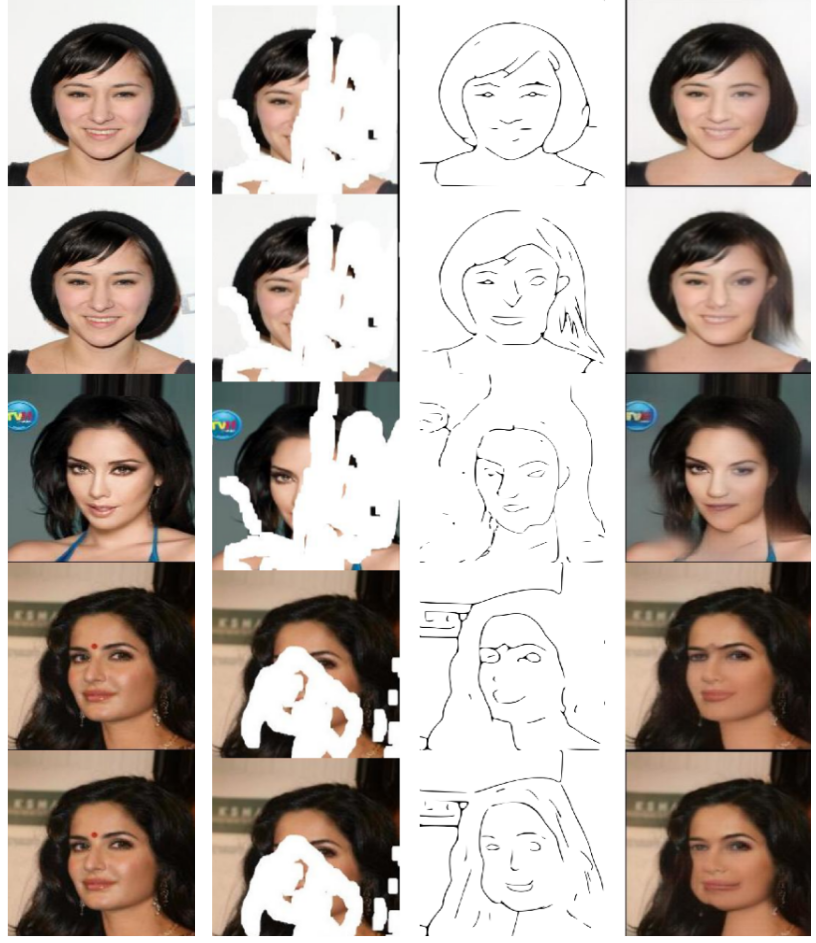}
\caption{Results obtained with sketch-based FIpoly-GAN. From left to right: original, damaged, sketch, inpainted image.}
\label{fig:skecth_based_method}
\end{figure}

Representative results generated by our sketch-based FIpoly-GAN method are shown in Figure \ref{fig:skecth_based_method}. The results 
% are a little blurry yet the 
illustrate that the network can generate realistic images by inpainting conditioned on a rough sketch.
% For comparison of our method with state of the art, we show results obtained with SC-FEGAN in Figure \ref{fig:SCF_examples}, where we used the same subject image and same sketch. We can see that our method significantly outperforms SC-FEGAN from results shown in figure \ref{fig:skecth_based_method} \ref{fig:SCF_examples}.

% \begin{figure}[h!]
%   \centering
%   \includegraphics[width=\columnwidth]{inpainting/SCF_examples.png}
% \caption{Results obtained with SC-FEGAN \cite{scfegan}. From left to right: original, damaged, damaged with sketch, inpainted.}
% \label{fig:SCF_examples}
% \end{figure}

\section{Conclusion}
% In this chapter we show the importance of edges in image recovery, inpainting and image editing. We compare all state of the art methods from the year 2018 to 2019, and on comparison we perform equally good or better than the state of the art for the same given task. Our designed method is suitable in case of availability of less data, less training resources, and our methods generalizes to random images very well.
In this paper, we presented FIpoly-GAN, a conditional GAN for extreme face inpainting.  Our work demonstrated the importance of edge information in the recovery of badly damaged regions for image restoration or editing. 
The FIpoly-GAN approach performs equally well or better than state-of-the-art methods on standard datasets and outperforms other methods on images 
% obtained from other sources
unrelated to the training data.
% than many current state-of-the-art for the same given task. 
% Our designed method is suitable for better generalization in cases where smaller datasets are used for training, and fewer training resources are available. 
% Our Poly-GAN based methods generalize  very well for images that are outside the training dataset.\cite{GAN}

\section{Acknowledgments}

This research was supported in part by the Air Force Office of Scientific Research (AFOSR) under Dynamic Data Driven
Applications Systems (DDDAS) grant FA9550-18-1-0121 and the Center for Emerging and Innovative Sciences
(CEIS), an Empire State Development-designated Center for Advanced
Technology..

%%%%%%%%%%%%%%%%%%%%%%%%%%%%%%%%%%
% Biography
%%%%%%%%%%%%%%%%%%%%%%%%%%%%%%%%%%

% \begin{biography}
% Please submit a brief biographical sketch of no more than 75 words. 
% Include relevant professional and educational information as shown 
% in the example below.

% Jane Doe received her BS in physics from the University of Nevada (1977) 
% and her PhD in applied physics from Columbia University (1983). Since 
% then she has worked in the Research and Technology Division at Xerox 
% in Webster, NY. Her work has focused on the development of toner adhesion 
% and transport issues. She is on the Board of  IS\&T and a member of APS 
% and SPIE.
% \end{biography}

%%%%%%%%%%%%%%%%%%%%%%%%%%%%%%%%%%
% Biography
%%%%%%%%%%%%%%%%%%%%%%%%%%%%%%%%%%

\begin{biography}

Nilesh Pandey completed his Bachelor in Electrical Engineering at Ramrao Adik Institute of Technology (2016) and received his Master in Computer Engineering from Rochester Institute of Technology (2019).
His research interests include Generative Adversarial Networks (GAN), and applications for tracking and pose estimation architectures.

Andreas Savakis is Professor of Computer Engineering and
Director of the Center for Human-aware Artificial Intelligence (CHAI)
at Rochester Institute of Technology (RIT). He received his Ph.D. in Electrical and Computer Engineering from North Carolina State University. Prior to joining RIT, he was Senior Research Scientist at Kodak Research
Labs. 
His research interests
include computer vision, deep learning, machine learning, domain adaptation, object tracking, human pose estimation, and scene analysis. 
Dr. Savakis has coauthored over 120 publications and is co-inventor on 12 U.S.
patents. 
\end{biography}
\end{document}